\documentclass[11pt]{article}
\usepackage{acl}

\usepackage{times}
\usepackage{latexsym}
\usepackage[T1]{fontenc}
\usepackage[utf8]{inputenc}
\usepackage{microtype}
\usepackage{inconsolata}
\usepackage{graphicx}

\usepackage{booktabs}
\usepackage{array}
\usepackage{tabularx}
\usepackage[table]{xcolor}
\usepackage{wrapfig}
\usepackage{multirow}

\definecolor{tabours}{HTML}{E8F0FA}
\definecolor{tabgroup}{HTML}{F4F6F8}
\definecolor{tabgain}{HTML}{2E7D32}
\definecolor{tabmuted}{HTML}{6B7280}
\definecolor{tabloss}{HTML}{C62828}
\definecolor{tabevid}{HTML}{1F5AA6}
\newlength{\gainpad}
\newcommand{\gain}[1]{\rlap{\,\scriptsize\textcolor{tabgain}{+#1}}}
\newcommand{\venue}[1]{\textcolor{tabmuted}{#1}}
\usepackage{pifont}
\newcommand{\cmark}{\textcolor{tabgain}{\ding{51}}}
\newcommand{\xmark}{\textcolor{tabloss}{\ding{55}}}
\newcommand{\tabsetup}{%
  \small
  \setlength{\tabcolsep}{4pt}%
  \setlength{\aboverulesep}{0pt}%
  \setlength{\belowrulesep}{0pt}%
  \renewcommand{\arraystretch}{0.92}%
  \setlength{\extrarowheight}{1.8pt}%
}
\newcolumntype{Y}{>{\centering\arraybackslash}X}
\newcolumntype{R}{>{\raggedleft\arraybackslash}X}
\newcolumntype{G}{>{\centering\arraybackslash}X<{\hspace*{\gainpad}}}
\usepackage{multicol}
\usepackage{amssymb}
\usepackage{amsmath}
\usepackage[most]{tcolorbox}
\tcbuselibrary{skins}
\usepackage{enumitem}

\newtcblisting{prompt}[1][]{
    title=\texttt{#1},
    listing only,
    listing options={style=mypython},
    colback=darkgray,
    colframe=gray!60,
    colbacktitle=gray!30,
    coltitle=black,
    fonttitle=\bfseries,
    enhanced,
    frame hidden=false,
    boxrule=0.5pt,
    arc=1mm,
    left=2mm,
    right=2mm,
    top=1mm,
    bottom=1mm
}

\usepackage{amsthm}
\usepackage{amsmath}
\usepackage{amssymb}

\title{From Documents to Spans: Span-Centric Learning for LLM-based Explainable Medical Coding}

\author{
 \textbf{Xu Zhang}\textsuperscript{1,2,5} ,
 \textbf{Wenxin Ma}\textsuperscript{1,2} ,
 \textbf{Chenxu Wu}\textsuperscript{\textbf{1,2}} ,
 \textbf{Rongsheng Wang}\textsuperscript{\textbf{1,2}} \textbf{,}\\
 \textbf{Zhiyang He}\textsuperscript{5} ,
 \textbf{Xiaodong Tao}\textsuperscript{5} ,
 \textbf{Kun Zhang}\textsuperscript{\textbf{1,2}}\footnotemark[1] \textbf{,}
 \textbf{S. Kevin Zhou}
 \textsuperscript{\textbf{1,2,3,4}}\footnotemark[1]\\
 \textsuperscript{1} School of Biomedical Engineering, Division of Life Sciences and Medicine, USTC\\
 \textsuperscript{2} MIRACLE Center, Suzhou Institute for Advance Research, USTC
\\
 \textsuperscript{3}Jiangsu Provincial Key Laboratory of Multimodal Digital Twin Technology
\\
 \textsuperscript{4}State Key Laboratory of Precision and Intelligent Chemistry, USTC
\\
  \textsuperscript{5}iFlyHealth
\\
{\tt\small xu\_zhang@mail.ustc.edu.cn kkzhang@ustc.edu.cn skevinzhou@ustc.edu.cn}
}

\begin{document}
\maketitle

\begin{abstract}
Medical coding, which assigns standardized ICD codes to clinical documents, is a fundamental yet challenging task in modern healthcare. It requires extensive knowledge of a large code space and the ability to identify and interpret sparse supporting evidence in long, noisy clinical documents.
Most existing methods rely on document-level supervision, which provides limited guidance on supporting evidence and insufficient training signals for rare codes.
To address these limitations, we propose Span-Centric Learning (SCL), which separates coding knowledge acquisition from its application in full documents.
SCL first learns coding knowledge from short spans, then learns to use this knowledge in full documents from a small set of annotated data. 
Experiments show that, our approach enables an 8B LLM to surpass state-of-the-art methods using only 248 complete documents for training.
It also provides explicit supporting evidence for each predicted code, with strong alignment to human annotations.
Further analysis shows that the model can learn knowledge of codes absent from documents through spans and apply that knowledge to full documents.
\end{abstract}

\section{Introduction}
\label{intro}

Medical coding assigns International Classification of Diseases (ICD) codes to clinical documents, supporting insurance reimbursement and health data analysis~\cite{www}. Accurate coding is challenging because clinical documents are long and noisy, while the ICD inventory contains thousands of fine-grained codes~\cite{problems}.  These challenges can lead to errors even among human coders~\cite{accuracy}. Beyond accurate code assignment, automated systems should provide supporting evidence that allows humans to inspect and review their predictions~\cite{mdace,plmca,less,icdagent}.

\begin{figure}[t]
    \centering
    \includegraphics[width=\linewidth]{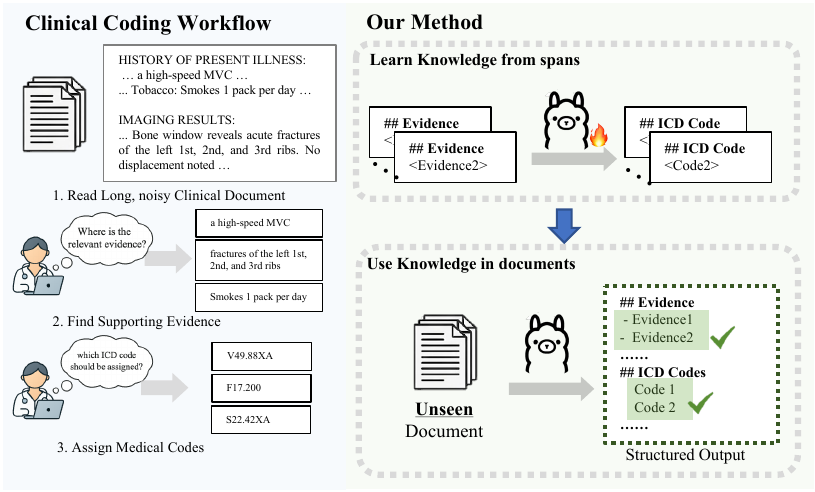}
    \vspace{-2mm}
    \caption{\textbf{Motivation.} Human coders assign medical codes by identifying supporting evidence in clinical documents. Motivated by this process, our method first learns broad coding knowledge from span data, and then transfers learned knowledge to document-level explainable medical coding.
    }
    \label{fig:teaser}
    \vspace{-2mm}
\end{figure}

Most supervised medical coding methods rely on full documents paired with code labels~\cite{caml,plmicd,icdagent}, which have two limitations. First, code coverage is uneven: even large datasets such as MIMIC-III and MIMIC-IV~\cite{mimiciii,mimiciv} exhibit long-tailed code distributions, leaving many codes with few or no training examples~\cite{align-icd,gki-icd,clh}. Second, document-level labels specify coding outcomes but not the passages supporting each assignment. Models must therefore infer evidence-to-code relationships within long, noisy documents, without explicit guidance on evidence selection for accurate coding and human review.

Addressing these limitations by scaling document-level supervision is costly.
ICD codes follow a long-tailed distribution in real-world clinical data, requiring large document collections to obtain sufficient examples of rare codes.
Moreover, annotating supporting evidence is labor-intensive, as expert coders must review long, noisy documents to identify the passages supporting each code assignment.
To the best of our knowledge, MDACE is the only publicly available medical coding dataset with expert-annotated supporting evidence, yet it contains only a few hundred encounters and covers a limited subset of ICD codes~\cite{mdace}.

In this work, we argue that \textit{medical coding need not be learned entirely from full documents}. The task requires both acquiring coding knowledge and learning to apply it in context. Much of this knowledge, including which clinical evidence warrants coding and how codes differ or relate, can be learned from short evidence spans. Learning this knowledge from spans and using only a small set of annotated documents to transfer it to full-document coding can therefore substantially reduce the need for costly document-level supervision.

Guided by this insight, we propose \textbf{Span-Centric Learning (SCL)}, a two-stage training framework for medical coding. 
In the first stage, the model learns coding knowledge from explicit evidence-to-code mappings in a large span library. 
In the second stage, a small set of evidence-annotated documents teaches the model to identify supporting evidence and assign codes in complex clinical contexts. Span replay provides continued evidence-to-code supervision, helping retain the acquired knowledge and guide its transfer to full-document coding.

To support this training strategy, we construct a span library with broad code coverage and reliable evidence-to-code mappings. We begin by extracting evidence spans from existing code-labeled documents. To extend coverage to codes with few or no extracted spans, we introduce \textit{Dynamic Context-Augmented Span Synthesis}, which generates additional spans using authoritative coding resources and related examples. To ensure the quality of these synthetic spans, we incorporate \textit{Iterative Refinement} into the synthesis process: generated spans are assessed, and feedback on rejected candidates guides subsequent rounds of generation. Together, these steps expand span-level supervision without requiring evidence-annotated documents covering every code.

We evaluate SCL on two widely used ICD coding benchmarks, MDACE~\cite{mdace} and ACI~\cite{verify}. 
An 8B model trained with SCL substantially outperforms task-specific agentic methods based on GPT-5, while supporting local deployment. 
SCL also pairs code predictions with explicit supporting evidence, facilitating human review. 
More importantly, SCL achieves significantly better coding performance and greater interpretability with less than 1\% full documents for training, highlighting the effectiveness and efficiency of span-centric learning.

To summarize, our contributions are threefold:
\begin{itemize}
    \item We present an intuitive decomposition of medical coding, showing that evidence spans can serve as the primary unit for acquiring coding knowledge, challenging prevailing paradigm of learning from large-scale full documents.
    \item We propose SCL through complementary training and data strategies. For training, a two-stage framework combines span-level pretraining with document-level adaptation and span replay. For data, targeted span synthesis and iterative refinement build a span library with broad code coverage and reliable evidence-to-code mappings.
    \item We enable an 8B LLM to achieve state-of-the-art medical coding performance using only 248 documents for full-document training, surpassing task-specific agentic method based on GPT-5 while producing supporting evidence closely aligned with human annotations.
\end{itemize}

\section{Related Work}

\textbf{Discriminative medical coding methods}. 
Early automated medical coding systems model this task as multi-label classification.
A representative paradigm is label attention~\cite{caml,plmicd,plmca}, where each ICD code is modeled by a learnable query vector that attends to the clinical text and independently determines whether the corresponding code should be assigned.
Subsequent work extends this framework by incorporating external knowledge~\cite{dkec,corelation,msmn,msam,gki-icd}.
Beyond knowledge enhancement, some studies seek to explain the predictions of black-box models~\cite{plmca,autocodedl}, while ~\citet{less} improve training efficiency through compact clinical-entity representations.
Despite these advances, discriminative methods still rely heavily on document-level supervision, where scarce evidence annotations and uneven code distributions are costly to address through additional annotation.

\vspace{0.5em}
\noindent
\textbf{LLM-based medical coding methods.}
LLMs can generate supporting evidence alongside code predictions, facilitating human review and correction.
Training-free approaches~\cite{llmicd,mac,medcoder,clh,travel} often combine multi-stage workflows with external knowledge bases, relying on powerful LLMs whose computational demands or proprietary access can hinder local deployment.
Post-training methods~\cite{verify,russian} improve the coding performance of smaller, locally deployable models.
However, existing datasets typically lack evidence annotations, making it challenging to preserve LLMs' inherent interpretability.
Recent work~\cite{icdagent} constructs rationale-augmented data for 15,000 MIMIC-IV notes to train a coding agent, but manual quality control of such document-level annotations remains challenging.
Consequently, how to train LLMs for explainable medical coding under realistic annotation constraints remains largely underexplored.

\section{Method}

\subsection{Overview}

\textbf{Problem formulation.}
Given a clinical document \(x\), the model predicts ICD codes \(C=\{c_j\}_{j=1}^{n}\) through supporting evidence spans \(E=\{e_i\}_{i=1}^{m}\):
\begin{equation}
\label{eq:problem}
x \rightarrow E \rightarrow C,
\end{equation}

\textbf{Existing paradigms.}
Traditional methods \cite{plmicd} simplify the reasoning process into a direct map \(x \rightarrow C\), sacrificing explainability.
Recent work~\cite{icdagent} constructs CoT data on such large-scale datasets, but the CoT data is difficult to maintain.

\textbf{Our solution.}
We decouple the process into a small set of high-quality CoT documents and a large library of maintainable spans.
Our goal is to first inject the dense knowledge needed for this task, and use a small set of CoT data to transfer knowledge to document-level reasoning.
The training objective mixes the two sources:
\begin{equation}
\label{eq:mixed_sft}
\begin{aligned}
\min_{\theta}\quad
&\mathbb{E}_{(x,y)\sim\mathcal{D}_{\mathrm{doc}}^{*}}
\left[
\mathcal{L}_{\mathrm{doc}}
\big(f_\theta(x), y\big)
\right] \\
&+
\mathbb{E}_{(e,c)\sim\mathcal{D}_{\mathrm{span}}}
\left[
\mathcal{L}_{\mathrm{span}}
\big(f_\theta(e), c\big)
\right],
\end{aligned}
\end{equation}
where $y=\mathrm{CoT}(E,C)$ is a high-quality reasoning trace constructed from the ground-truth codes $C$ and their supporting evidence $E$.
$\mathcal{D}_{\mathrm{doc}}^{*}$ is a small and maintainable set of such documents.
$\mathcal{D}_{\mathrm{span}}$ is scalable, maintainable, and can be expanded from multiple sources.

\begin{figure*}[t]
  \includegraphics[width=\linewidth]{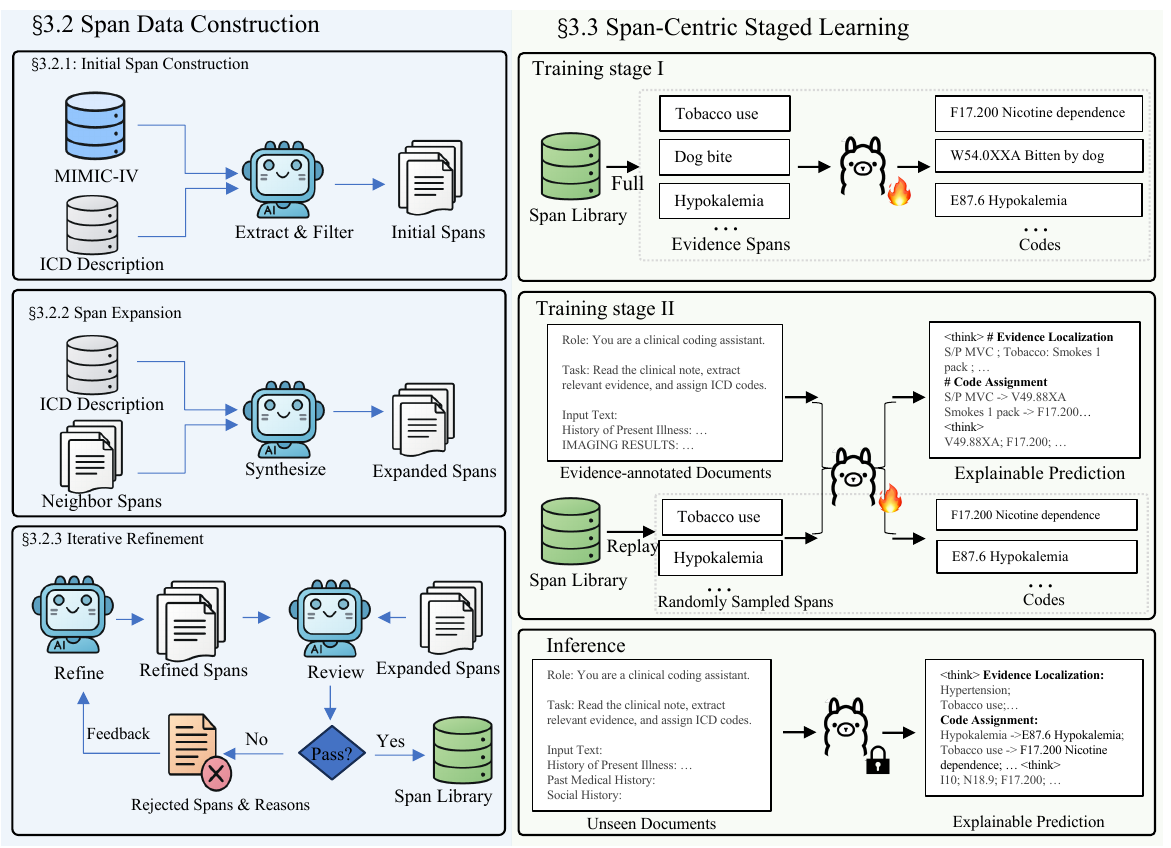}
  \vspace{-1.5mm}
  \caption{\textbf{Overview of SCL framework.}
   Span Data Construction (Sec \ref{sec:span_construction}) constructs, expands and refines a span library from public medical coding datasets.
   Span-Centric Staged Learning (Sec \ref{sec:staged_training}) first learns broad coding knowledge from the span library, then learns to apply knowledge to document-level medical coding from a small set of high-quality CoT data, with mixed-batch replay of a randomly sampled fraction of the span library. At inference, the model identifies evidence spans in unseen documents and assigns corresponding codes.
  }
  \label{fig:overview}
  \vspace{-1.5mm}
\end{figure*}

As shown in Fig \ref{fig:overview}, we first present \textit{Span Data Construction} in Sec.~\ref{sec:span_construction}: we construct an initial span library, expand it via \textit{Dynamic Context-Augmented Span Synthesis} in Sec.~\ref{SCL_data_expansion}, and control its quality via \textit{Iterative Refinement} in Sec.~\ref{sec:iterative_refinement}.
Using this library, \textit{Span-Centric Staged Training} in Sec.~\ref{sec:staged_training} first injects span knowledge and then uses it to enhance document-level reasoning over the small document dataset.

\subsection{Span Data Construction}
\label{sec:span_construction}

\subsubsection{Initial Span Data Construction}
\label{sec:initial_span}

We construct $\mathcal{D}_{\mathrm{span}}$ from public medical coding datasets that provide document-level code labels but no evidence annotations.
The construction pipeline comprises extraction, aggregation, and augmentation.

\textbf{Extract.}
Given a labeled document sample $(x,C)$, for each code $c\in C$, we prompt Llama-3.3-70B to extract an evidence span $e\subseteq x$ that directly supports $c$.
The model may return ``No Evidence Found'' when the assigned code cannot be substantiated by the document, preventing forced evidence generation.

\textbf{Aggregate.}
We normalize and aggregate the extracted spans by code, retaining unique spans and their occurrence frequencies.
We discard invalid spans like ``No Evidence Found''.
To limit the dominance of frequent codes, for each code, we retain the top $B=10$ most frequent spans.

\textbf{Augment.}
Finally, we supplement the mined pairs with two authoritative lexical sources: terms from the ICD-10 Alphabetic Index paired with their default codes, and official code descriptions from the ICD-10 Tabular List.
These sources expand coverage beyond the codes reliably recovered from public documents.

\subsubsection{Dynamic Context-Augmented Span Synthesis}
\label{SCL_data_expansion}

Rare codes on the long tail lack qualified spans.
Inspired by retrieval-augmented generation~\cite{rag}, we synthesize new candidate spans $\tilde{E}(c)$ from the official description $d(c)$ and a dual context,
\begin{equation}
\tilde{E}(c)=g\big(
\underbrace{d(c)}_{\mathrm{query}},\,
\underbrace{R^{+}(c),\, R^{-}(c)}_{\mathrm{augmented}}
\big),
\label{eq:dual_context}
\end{equation}
where $R^{+}(c)$ and $R^{-}(c)$ are the dual context of code $c$.
This context is dynamic and changes with the number of synthesis rounds.

\textbf{Neighbor span augmentation.}
Qualified spans of neighbor codes are
\begin{equation}
R^{+}(c)=\{(e^{+},c')\mid c'\in\mathcal{N}(c),\, e^{+}\in Q(c')\}.
\label{eq:pos_context}
\end{equation}
$\mathcal{N}(c)$ contains hierarchically adjacent codes, such as parents and siblings, and $Q(c')$ holds the spans of $c'$ that have passed assessment.
Adjacent codes share clinical wording and differ by a few attributes, such as anatomy, laterality, acuity, and etiology.
Llama-3.3-70B takes such a span as a scaffold and rewrites these attributes to match $d(c)$.

\textbf{Rejected span augmentation.}
Unqualified spans of the target code are
\begin{equation}
R^{-}(c)=\{(e^{-},r)\mid e^{-}\in\bar{Q}(c)\}.
\label{eq:neg_context}
\end{equation}
$\bar{Q}(c)$ holds the spans of $c$ rejected so far, and $r$ states whether $e^{-}$ failed naturalness or plausibility.
The generator uses $r$ to avoid the same defect: it removes non-clinical phrasing and restores missing distinguishing attributes.

Synthesized spans return to the loop as candidates (Sec.~\ref{sec:iterative_refinement}).
$R^{-}(c)$ is empty at the start.
As the number of rounds grows, qualified spans of adjacent codes accumulate in $R^{+}(c)$, and rejected spans with their reasons accumulate in $R^{-}(c)$.

\subsubsection{Iterative Refinement}
\label{sec:iterative_refinement}

We refine the span library with a closed loop of assessment and expansion to ensure that all spans retained for each code satisfy the quality criteria.

\textbf{Criteria.}
A candidate must satisfy two criteria: naturalness and plausibility.
\textit{Naturalness} requires an expression that would appear in a clinical document, rather than a non-clinical, contradictory, or unseen phrasing.
\textit{Plausibility} requires the span to support its assigned code, including attributes that distinguish the code from its neighbors, such as anatomy, acuity, laterality, and etiology.
The same criteria are used for the human evaluation in Appendix~\ref{sec:appendix_span_quality}.

\textbf{Loop.}
Let $Q(c)$ be the spans of code $c$ that pass both criteria.
A candidate is admitted into $Q(c)$ only when it meets the criteria above; otherwise it is stored with its rejection reason.
For each code with unqualified candidates, Llama-3.3-70B generates replacement candidates from $d(c)$ and the dual context in Eq.~\eqref{eq:dual_context}, and these candidates return to assessment.
The loop stops when all retained spans satisfy both criteria.

\subsection{Span-Centric Staged Learning}
\label{sec:staged_training}

Using the span library constructed above, we divide the training process into two stages.
The model learns local coding knowledge from spans at the first stage, and then uses that knowledge to enhance document-level reasoning at the second stage.

\textbf{Stage I: knowledge injection.}
Let $\mathcal{D}_{\mathrm{span}}=\{(e,c)\}$ denote the scalable span corpus.
Given an evidence span $e$ from $\mathcal{D}_{\mathrm{span}}$, the model is trained to generate its ICD code $c$:
\begin{equation}
\theta_{1}=\arg\min_{\theta}
\mathbb{E}_{(e,c)\sim\mathcal{D}_{\mathrm{span}}}
\left[-\log p_{\theta}(c\mid e)\right].
\label{eq:stage_one}
\end{equation}
This local mapping teaches the model not only the evidence-to-code mapping, but also the expressions of codeable evidence, which helps localize evidence in long documents.
Because $e$ is substantially shorter than a full clinical document, broad code coverage can be obtained with fewer training tokens.

\textbf{Stage II: enhance reasoning with knowledge.}
Document-level reasoning traces are scarce in existing ICD coding datasets.
MIMIC-IV lacks evidence annotations and contains unreliable code labels (Appendix~\ref{sec:appendix_benchmarks}).
We therefore construct high-quality CoT data $\mathcal{D}_{\mathrm{doc}}^{*}$ using evidence-annotated documents from MDACE and documents from ACI augmented with LLM-generated evidence annotations.

Each document is converted into an evidence-grounded reasoning trace, writing the supporting evidence before assigning the corresponding code(s):
\begin{equation}
y(x)=\bigoplus_{i=1}^{K}\left[e_i\,\Vert\,c_i\,\Vert\,d(c_i)\right],
\label{eq:structured_target}
\end{equation}
where $e_i$ is the $i$-th of $K$ evidence spans in document order, $c_i$ its corresponding codes, and $d(c_i)$ their official descriptions.
Both $\Vert$ and $\bigoplus$ denote concatenation.
This supervision teaches the model to follow the human coding workflow, grounding each coding decision in its supporting evidence.

Starting from $\theta_{1}$, we enhance this document-level reasoning with the span knowledge acquired in Stage I:
\begin{equation}
\begin{aligned}
\mathcal{L}_{\mathrm{II}}(\theta)=
&\;\mathbb{E}_{(x,y)\sim\mathcal{D}_{\mathrm{doc}}^{*}}
  \left[-\log p_{\theta}(y\mid x)\right]\\
&+\lambda\,
\mathbb{E}_{(e,c)\sim\mathcal{D}_{\mathrm{span}}}
  \left[-\log p_{\theta}(c\mid e)\right],
\end{aligned}
\label{eq:stage_two}
\end{equation}
Training exclusively on the small document set may overwrite the broad code knowledge acquired in Stage I.
We therefore mix document examples with randomly sampled Stage-I examples in every batch.
Here $\lambda$ weights the replay sample ratio, i.e., how many Stage-I examples enter each batch.
\section{Experiments} \label{Experiments}

\subsection{Experimental Setup}
\label{setup}

\textbf{MIMIC-IV} \cite{mimiciv} is currently the largest publicly available dataset annotated with ICD-10 codes. However, MIMIC-IV only contains discharge summaries, while some ICD codes are assigned based on other clinical notes that are not included in the dataset. As a result, many codes are not supported by the available text \cite{mdace, survey_replication,verify}, making MIMIC-IV unsuitable as a reliable benchmark (Appendix \ref{sec:appendix_benchmarks}). We therefore treat it as a large but noise-prone training dataset, and rely on high-quality external benchmarks to measure true coding performance.

\textbf{MDACE} \cite{mdace} is an expert-annotated subset of MIMIC-III \cite{mimiciii}, containing gold-standard evidence span annotations. It has no overlap with the MIMIC-IV ICD-10 subset and has become increasingly popular as a benchmark for evaluating both ICD coding accuracy and evidence faithfulness.

\textbf{ACI} \cite{aci-bench} is an outpatient dataset originally developed for AI scribes, based on which \citet{verify} construct a new double expert-annotated ICD-10-CM coding benchmark.

We extract spans from MIMIC-IV and use documents from the downstream training sets.
The baselines are directly fine-tuned on the union of MIMIC-IV, MDACE, and ACI.
For discriminative models, we use the validated-optimal threshold.
For LLMs, we extract the ICD code from the text output.

\subsection{Comparison with SOTA Methods}
\label{compare}

\begin{table*}[t]
    \centering
    \caption{\textbf{Performance comparison with state-of-the-art methods.}
    F1, P, and R denote micro-averaged F1, precision, and recall; Avg is the mean F1 over the two benchmarks.
    \textbf{Bold}: best in each setting; \textcolor{tabgain}{green}: gain over the best baseline.}
    \label{table:compare}
    \small
    \setlength{\tabcolsep}{4pt}
    \setlength{\aboverulesep}{0pt}
    \setlength{\belowrulesep}{0pt}
    \renewcommand{\arraystretch}{0.92}
    \setlength{\extrarowheight}{1.8pt}
    \newcolumntype{F}{>{\centering\arraybackslash}X<{\hspace*{\gainpad}}}
    \begin{tabularx}{\textwidth}{l l FFF FFF F}
    \toprule
    \multirow{2}{*}{\textbf{Method}} & \multirow{2}{*}{\textbf{Venue}}
    & \multicolumn{3}{c}{\textbf{ACI}\hspace*{\gainpad}} & \multicolumn{3}{c}{\textbf{MDACE}\hspace*{\gainpad}} & \textbf{Avg} \\
    \cmidrule(lr{\dimexpr\gainpad+\cmidrulekern\relax}){3-5}
    \cmidrule(lr{\dimexpr\gainpad+\cmidrulekern\relax}){6-8}
    \cmidrule(lr{\dimexpr\gainpad+\cmidrulekern\relax}){9-9}
    & & F1 & P & R & F1 & P & R & F1 \\
    \midrule
    \rowcolor{tabgroup}
    \multicolumn{9}{l}{\textbf{\textit{Full codes}} \textcolor{tabmuted}{(complete ICD-10-CM code space, ${\sim}$70k codes)}} \\
    PLM-ICD       & \venue{NAACL 22 WS}  & 21.0 & 12.5 & 65.2 & 54.8 & 52.4 & \textbf{57.5} & 37.9 \\
    GKI-ICD       & \venue{ACL'25}   & 42.6 & 31.8 & 64.4 & 54.5 & 53.5 & 55.5 & 48.6 \\
    MedCodER-GPT5 & \venue{NAACL'25} & 62.0 & 57.0 & 68.0 & 40.0 & 37.0 & 42.0 & 51.0 \\
    CLH-GPT5      & \venue{EMNLP'25} & 57.2 & 58.5 & 56.0 & 46.0 & 40.0 & 55.0 & 51.6 \\
    ICDAgent      & \venue{ACL'26}   & 46.6 & 41.3 & 53.4 & 50.9 & 47.4 & 55.0 & 48.8 \\
    \rowcolor{tabours}
    \textbf{SCL-Llama3-8B (Ours)} & & \textbf{67.3}\gain{5.3} & \textbf{62.2}\gain{3.7} & \textbf{73.3}\gain{5.3} & \textbf{57.2}\gain{2.4} & \textbf{57.3}\gain{3.8} & 57.1 & \textbf{62.3}\gain{10.7} \\
    \midrule
    \rowcolor{tabgroup}
    \multicolumn{9}{l}{\textbf{\textit{Restricted codes}} \textcolor{tabmuted}{(benchmark code space, ${\sim}$1k codes)}} \\
    MedDCR-GPT5 & \venue{ACL'26}   & 52.0 & 43.0 & 67.0 & 51.0 & 46.0 & \textbf{59.0} & 51.5 \\
    \rowcolor{tabours}
    \textbf{SCL-Llama3-8B (Ours)} & & \textbf{74.0}\gain{22.0} & \textbf{74.8}\gain{31.8} & \textbf{73.3}\gain{6.3} & \textbf{59.4}\gain{8.4} & \textbf{61.8}\gain{15.8} & 57.1 & \textbf{66.7}\gain{15.2} \\
    \bottomrule
    \end{tabularx}
    \end{table*}

\begin{table*}[t]
  \centering
 \caption{\textbf{Full-document supervision vs. span-centric supervision.}
All models use Llama3-8B and are trained on a single H20 GPU.
SFT and SFT\textsuperscript{E} use about 120k training documents from MIMIC-IV, MDACE, and ACI, with the latter augmented with evidence generated by Llama-3.3-70B for document-level distillation.
SCL uses 248 evidence-annotated documents and 200k spans.
$^{*}$ indicates $p<10^{-2}$ vs. SFT (two-sided paired permutation tests, Holm-adjusted).
  \textcolor{tabgain}{Green}: gain over SFT.}
  \label{table:span_vs_document}
  \tabsetup
  \begin{tabularx}{\textwidth}{l cc YY YY c c}
  \toprule
  \multirow{2}{*}{\textbf{Method}}
  & \multicolumn{2}{c}{\textbf{Training data}}
  & \multicolumn{2}{c}{\textbf{ACI}}
  & \multicolumn{2}{c}{\textbf{MDACE}}
  & \multirow{2}{*}{\textbf{Time (hr)}}
  & \multirow{2}{*}{\textbf{Explainability}} \\
  \cmidrule(lr){2-3} \cmidrule(lr){4-5} \cmidrule(lr){6-7}
  & Docs & Spans & Micro-F1 & Macro-F1 & Micro-F1 & Macro-F1 & & \\
  \midrule
  SFT
  & 120k & \textcolor{tabmuted}{--}
  & 47.9 & 15.7 & 47.0 & 24.0
  & 252 & \xmark \\
  SFT\textsuperscript{E}
  & 120k & \textcolor{tabmuted}{--}
  & 53.6 & 29.6 & 55.7 & 30.2
  & 264 & \cmark \\
  \rowcolor{tabours}
  \textbf{SCL (Ours)}
  & 248 & 200k
  & \textbf{67.3}$^{*}$\gain{19.4} & \textbf{41.0}\gain{25.3}
  & \textbf{57.2}$^{*}$\gain{10.2} & \textbf{32.4}\gain{8.4}
  & \textbf{22} & \cmark \\
  \bottomrule
  \end{tabularx}
  \end{table*}

\noindent\textbf{Evaluation.} We report results under \textbf{full codes}, which scores predictions over the complete ICD-10-CM code space (about 70k codes). We adopt this setting because real-world deployment must handle all assignable codes, and therefore it more faithfully reflects practical coding performance.
To make comparison with prior work, we also report results under \textbf{restricted code evaluation}, which restricts the candidate code set to codes that appear in each benchmark.

\noindent\textbf{Baselines.} 
We compare against recent task-specific agentic methods, including MedCodER \cite{medcoder}, CLH \cite{clh}, and MedDCR \cite{MedDCR}, and use GPT-5 as the backbone. We also compare with ICDAgent \cite{icdagent}, an 8B post-training method, which consists of two models enhanced by supervised fine-tuning and reinforcement learning.
For traditional supervised methods, we include PLM-ICD \cite{plmicd} and its knowledge enhanced version GKI-ICD \cite{gki-icd}.

\noindent\textbf{Results.}
As shown in Table~\ref{table:compare}, full-code evaluation better reflects real-world coding by requiring predictions over the complete code space, yielding lower scores than restricted-code evaluation. SCL-Llama3.1-8B consistently achieves the best Micro-F1 on both benchmarks, outperforming the strongest baselines on ACI and MDACE by 5.3 and 2.4 points under full codes, and by 22.0 and 8.4 points under restricted codes, respectively. Traditional supervised methods remain competitive on MDACE but perform poorly on ACI, suggesting overfitting to MIMIC-IV, which dominates the training data. In contrast, LLM-based methods show stronger cross-domain generalization, likely benefiting from their pretrained knowledge.

\subsection{Full Document Supervision vs. Span-centric Supervision}
\label{sec:data-scale-comparison}

\noindent\textbf{Setting.}
Table~\ref{table:span_vs_document} compares SCL with two document-level SFT baselines under the same backbone. Both baselines use all training documents from MIMIC-IV, MDACE, and ACI (approximately 120k documents in total).
Standard SFT learns to predict ICD codes from full documents, providing a reference for evaluating the training efficiency of span-centric supervision.
SFT\textsuperscript{E} additionally uses evidence generated by Llama-3.3-70B for these documents, serving as a document-level knowledge distillation baseline to assess the benefit of injecting coding knowledge at the span level.
In contrast, SCL injects coding knowledge through 200k short spans and uses only 248 evidence-annotated documents from the MDACE and ACI training sets for document-level training.

\noindent\textbf{Results.}
Compared with standard SFT, SCL improves Micro-F1 from 47.0 to 57.2 on MDACE and from 47.9 to 67.3 on ACI, with corresponding Macro-F1 gains of 8.4 and 25.3 points, while reducing training time from 252 to 22 hours (an 11.5$\times$ speedup). This comparison demonstrates the efficiency of span-centric supervision while maintaining higher coding accuracy.
Compared with SFT\textsuperscript{E}, SCL achieves Micro-F1 gains of 1.5 points on MDACE and 13.7 points on ACI, supporting the advantage of span-level knowledge injection over document-level distillation in this setting.

\begin{table}[t]
\centering
\caption{\textbf{Evaluation of model-predicted evidence on MDACE.}
We additionally include LLM-based metrics, since there may be multiple semantically equivalent evidence spans for one code in a document.
All scores are percentages; \textbf{bold}: better; \textcolor{tabgain}{green}: gain over Doc.}
\label{table:interpretability}
\tabsetup
\begin{tabularx}{\linewidth}{l >{\hsize=0.9\hsize}Y >{\hsize=1.1\hsize\columncolor{tabours}}G}
\toprule
\textbf{Metric} & \textbf{Doc} & \textbf{Doc \& Span}\hspace*{-\gainpad} \\
\midrule
\rowcolor{tabgroup}
\multicolumn{3}{l}{\textbf{\textit{Rule-based}} \textcolor{tabmuted}{(lexical matching)}} \\
Exact Match Rate       & \textbf{99.3} & 98.9 \\
Hit Rate (IoU $>$ 0)   & 67.1 & \textbf{74.5}\gain{7.4} \\
Hit Rate (IoU $>$ 0.5) & 47.0 & \textbf{54.2}\gain{7.1} \\
\midrule
\rowcolor{tabgroup}
\multicolumn{3}{l}{\textbf{\textit{LLM-based}} \textcolor{tabmuted}{(semantic matching)}} \\
Recall & 64.6 & \textbf{73.9}\gain{9.2} \\
F1     & 67.4 & \textbf{71.9}\gain{4.5} \\
\bottomrule
\end{tabularx}
\end{table}

\subsection{Explainability}
\label{explainability}

We evaluate predicted evidence against human-annotated spans on MDACE using Exact Match Rate (verbatim occurrence in the source document) and Hit Rate (token-level IoU $>0$ and $>0.5$ with annotated spans). To capture semantic agreement beyond lexical overlap, we also use GPT-5.1 to establish one-to-one span matches and compute evidence-level Recall and F1. The full protocol and prompt are provided in Appendix~\ref{sec:interpretability_prompt}.

As shown in Table~\ref{table:interpretability}, generated evidence is highly traceable to the source text, with Exact Match Rates above 98\% for both settings; unmatched spans mainly reflect minor textual changes, such as spelling errors or abbreviation expansion (Appendix~\ref{sec:appendix_failure_case}). The evidence also aligns closely with human annotations: Doc \& Span achieves a 74.5\% Hit Rate at IoU $>0$ and an LLM-based F1 of 71.9\%. Span-centric learning further improves evidence capture, increasing Hit Rate from 47.0\% to 54.2\% at IoU $>0.5$ and semantic recall from 64.6\% to 73.9\%.

\subsection{Analysis}
\label{finding}

\begin{figure}[t]
    \centering
    \includegraphics[width=\linewidth]{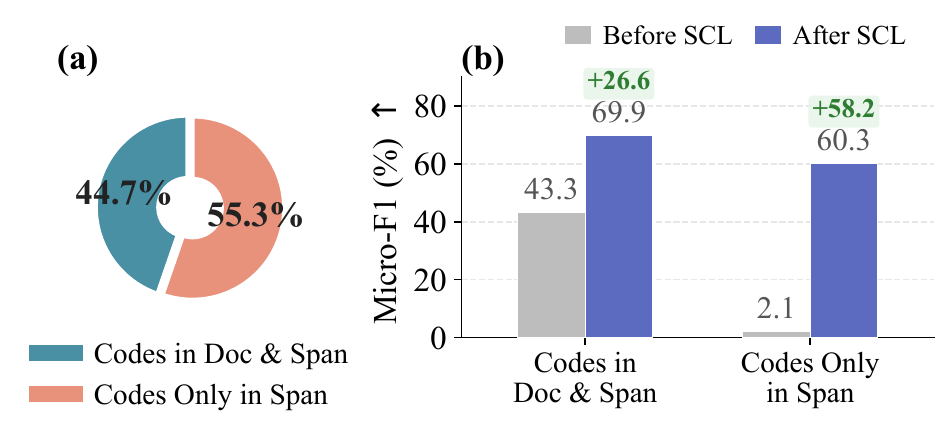}
    \caption{
        \textbf{Learning new code knowledge from spans on ACI.}
        (a) We divide ACI test-set codes into two groups: codes covered by both the 248 training documents and the span library, and codes covered by the span library but absent from all 248 training documents.
        (b) Span-level supervision substantially improves coding performance for codes absent from these documents, without requiring additional training documents.}
    \label{fig:effect-on-code}
\end{figure}

\subsubsection{Learning New Knowledge via Spans}
To examine whether span data can introduce code knowledge beyond that covered by the 248 training documents, we analyze ACI test-set codes that are absent from all of these documents but covered by the span library.
As illustrated in Figure~\ref{fig:effect-on-code}(a), we categorize ACI test-set codes into two groups: \textbf{codes in docs \& spans}, which are covered by both the 248 training documents and the span library, and \textbf{codes only in spans}, which are covered by the span library but absent from all 248 training documents.
We then compare coding performance on ACI before and after adding span data.

The ACI results in Figure~\ref{fig:effect-on-code}(b) show that SCL substantially improves coding accuracy for codes absent from all 248 training documents.
These gains indicate that span-level supervision can extend code knowledge beyond the coverage of the document-level training set.
The model can learn about these codes from span data and apply this knowledge to full-document coding without requiring additional training documents.

\subsubsection{Span Library Quality}
As shown in Table~\ref{table:span-refinement}, iterative refinement assesses candidate spans and replaces unqualified ones, improving naturalness from 96.50\% to 100.00\% and plausibility from 66.51\% to 100.00\% after synthesis.
Human and LLM evaluations of 4,645 final spans from 1,000 sampled codes agree on 97.57\% of naturalness judgments and 96.81\% of plausibility judgments (Appendix~\ref{sec:appendix_span_quality}), supporting LLM-based quality assessment.

\begin{table}[t]
\centering
\caption{\textbf{Span quality across iterative refinement.}
Naturalness and plausibility are pass rates (\%) under LLM-based assessment.}
\label{table:span-refinement}
\tabsetup
\begin{tabularx}{\linewidth}{l YY}
\toprule
\textbf{Span library} & \textbf{Naturalness} & \textbf{Plausibility} \\
\midrule
Initial spans      & 98.71 & 77.38 \\
+ Span Synthesis   & 96.50 & 66.51 \\
\rowcolor{tabours}
\textbf{+ Iterative Refinement} & \textbf{100.00} & \textbf{100.00} \\
\bottomrule
\end{tabularx}
\end{table}

\subsection{Ablation Study}
\label{ablation}

\begin{table}[t]
\centering
\caption{\textbf{Ablation of training strategies.} 
We compare different strategies for mixing long documents and short spans. 
Our two-stage strategy achieves the best performance while substantially shortening training time.}
\label{table:training-strategy}
\tabsetup
\begin{tabularx}{\linewidth}{l YY}
\toprule
\textbf{Strategy} & \textbf{Micro-F1} & \textbf{Time (hr)} \\
\midrule
\rowcolor{tabgroup}
\multicolumn{3}{l}{\textbf{\textit{One-stage}}} \\
Direct mixing               & 52.8 & 50 \\
Separate long/short batches & 42.9 & 20 \\
\midrule
\rowcolor{tabgroup}
\multicolumn{3}{l}{\textbf{\textit{Two-stage}} } \\
w/o replay & 43.8 & 22 \\
\rowcolor{tabours}
\textbf{w/ replay} & \textbf{57.2} & 22 \\
\bottomrule
\end{tabularx}
\end{table}

\noindent\textbf{Multi-stage training with mixed-batch replay.}
Our two-stage strategy improves Micro-F1 from 52.8 to 57.2 while reducing training time from 50 to 22 hours compared with one-stage mixing.
Replaying short spans alongside documents in Stage II is crucial: removing replay lowers Micro-F1 to 43.8, while placing spans and documents in separate batches yields only 42.9.
These results highlight the importance of multi-stage training and within-batch mixing for transferring span knowledge to document-level coding.

\begin{table}[t]
\centering
\caption{\textbf{Incremental ablation of span-centric supervision.}
Starting from initial spans, we progressively add span synthesis and iterative refinement.
Span synthesis uses no rejected spans or rejection reasons; this feedback is introduced with iterative refinement.}
\label{table:span-ablation}
\tabsetup
\begin{tabularx}{\linewidth}{l YY}
\toprule
\textbf{Variant} & \textbf{Micro-F1} & \textbf{Macro-F1} \\
\midrule
\rowcolor{tabgroup}
\multicolumn{3}{l}{\textbf{\textit{Documents only}}} \\
w/o evidence CoT & 36.8 & 13.9 \\
w/ evidence CoT  & 43.9 & 17.9 \\
\midrule
\rowcolor{tabgroup}
\multicolumn{3}{l}{\textbf{\textit{Documents \& spans}}} \\
Initial spans                  & 48.9 & 22.7 \\
+ Span Synthesis               & 53.7 & 27.9 \\
\quad w/o neighbor code spans  & 51.7 & 25.8 \\
\rowcolor{tabours}
\textbf{+ Iterative Refinement (Full)} & \textbf{57.2} & \textbf{32.4} \\
\bottomrule
\end{tabularx}
\end{table}

\noindent\textbf{Evidence CoT Guidance.}
With documents alone, adding evidence CoT raises Micro-F1 from 36.8 to 43.9 and Macro-F1 from 13.9 to 17.9, highlighting the importance of process supervision through evidence annotations for learning both evidence localization and accurate code assignment.

\noindent\textbf{Progressive span quality improvement.}
Adding initially extracted spans to the small document set improves Micro-F1 from 43.9 to 48.9 and Macro-F1 from 17.9 to 22.7, demonstrating that span-level coding knowledge transfers to document-level reasoning.
Span synthesis then expands coverage for under-covered long-tailed codes, raising Micro-F1 to 53.7 and Macro-F1 to 27.9 without rejection feedback; the neighbor-code context ablation evaluates this stage alone.
Finally, iterative refinement filters implausible spans and expands under-covered codes using rejected spans and rejection reasons as feedback, achieving 57.2 Micro-F1 and 32.4 Macro-F1.

\section{Discussion}
\label{sec:discussion}

\noindent\textbf{Rethinking Data Scaling.}
SCL challenges the reliance on massive datasets by surpassing previous methods using only 248 evidence-annotated documents, requiring just 3--5 spans per code for effective supervision. While the ``quality over quantity'' principle is recognized in general domains~\cite{lima,quality}, validating it within the complex, OOD-prone landscape of medical coding~\cite{rl} is meaningful, suggesting that performance depends less on raw data volume and more on high-quality, task-aligned annotations.

\noindent\textbf{Decomposing Clinical Complexity.}
By decoupling localized medical knowledge from document-level reasoning, SCL distills complex supervision into short, verifiable spans. This task decomposition dramatically lowers the barrier to acquiring expert medical annotations. Beyond medical coding, this strategy offers a scalable blueprint for tackling other complex, document-level clinical NLP tasks.
\section{Conclusion}
\label{sec:conclusion}

We propose Span-Centric Learning (SCL), a training framework for explainable medical coding that substantially reduces the need for full-document supervision.
SCL separates coding knowledge acquisition from its application in full documents, combining span-level learning with document-level adaptation and span replay.
Using only 248 documents for full-document training, SCL outperforms document-level fine-tuning baselines in coding accuracy and training efficiency, while producing explicit supporting evidence closely aligned with human annotations.
Further analysis shows that span-level supervision extends code knowledge beyond the coverage of these training documents and enables its application to full-document coding.

\section*{Limitations}

First, our span library is synthesized by LLM. Access to a human-annotated span library would help assess the upper bound of our method.
Second, our evaluation is limited to ICD-10-CM, leaving generalisation to other ontologies, such as ICD-10-PCS, CPT, and ICD-10-AM, unexplored.
Third, the current experiments are conducted on English clinical notes with ICD-10 codes.
While the proposed framework is language-agnostic in principle, we do not evaluate multilingual or cross-lingual settings.
Extending SCL to multilingual settings would be a valuable direction.

\section*{Acknowledgement}
This work is supported by the Natural Science Foundation of China under Grants 62271465 and 62502490; the National Key R\&D Program of China under Grant 2025YFC3408300; the Natural Science Foundation of Jiangsu Province under Grant BK20250496; the Suzhou Basic Research Program under Grant SYG202338; Jiangsu Funding Program for Excellent Postdoctoral Talent, and the China Postdoctoral Science Foundation under Grant 2024M763178.

\bibliography{custom}

\clearpage
\appendix

\section*{Appendix}
\label{sec:appendix}

\section{ICD Coding Background}
\label{sec:appendix_background}

\subsection{Task Definition}  
The ICD Coding task recognizes diseases, symptoms, conditions and procedures in a medical document, including discharge summaries, progress notes, and operative reports, and assigns standardized ICD codes to them. This task plays a critical role in healthcare administration, clinical statistics, reimbursement systems, and medical research.

From a computational perspective, ICD Coding is commonly formulated as a text-to-code prediction problem. Given a patient-level clinical document, the model is required to output a set of ICD codes. The task is characterized by a large label space, multi-label occurrence and severe label imbalance, which together make ICD Coding a challenging and distinctive problem in clinical natural language processing.

\subsection{Distinction from Related Tasks}

\textbf{Difference from diagnosis}.
Diagnosis aims to infer or determine what diseases a patient has, often involving clinical reasoning, uncertainty management, and causal inference. In contrast, ICD Coding does not seek to generate new diagnostic conclusions. 
Instead, it focuses on assigning standardized codes based solely on diagnoses and clinical facts that have already been documented by doctors. 
Therefore, ICD coding should be viewed as an information standardization task rather than a decision-making task.

\textbf{Difference from information extraction.}
ICD coding can be viewed as evidence extraction followed by code normalization, whereas NER and RE involve evidence extraction followed by type classification. 
Differently, ICD coding assigns each code only once, even if several spans support it, while NER densely annotates every mention.
Moreover, ICD coding involves a substantially larger label space with thousands of standardized codes, compared to the relatively small set of entity or relation types in NER and RE.

\subsection{ICD Coding Benchmarks}
\label{sec:appendix_benchmarks}

\textbf{Imperfect MIMIC data.}
Widely used datasets like MIMIC-III/IV suffer from annotation noise \cite{mdace,survey_replication,verify}.
The public MIMIC-IV release provides discharge summaries, while coders assign ICD codes from the full chart, including other documents that are not released.
Many codes therefore have no supporting evidence in the available text.

\textbf{MDACE.}
MDACE was introduced to supply expert evidence annotations for ICD-10 coding \cite{mdace}.
It re-annotates the ICD-9 portion of MIMIC-III with ICD-10.
MIMIC-IV overlaps MIMIC-III in time, but visits coded with ICD-9 and visits coded with ICD-10 are different encounters, so MDACE documents cannot overlap with the ICD-10 visits in MIMIC-IV.

\textbf{Remaining gaps.}
Although MDACE provides high-quality evidence annotations, its limited scale makes it insufficient as the sole source of training data.
MIMIC-IV, in turn, is difficult to free from annotation noise, and both datasets cover only a fraction of the full ICD ontology \cite{clh}.
Training on such noisy and truncated data risks forcing LLMs to overfit to dataset-specific biases rather than developing general medical coding capability.
Our span-centric paradigm offers a promising alternative to free from such annotation noise and dataset-specific biases.

\subsection{Authoritative Resources in ICD Coding}\label{kb}

\textbf{Alphabetic Index}. The Alphabetic Index maps clinical terms and their lexical variants to candidate ICD codes, bridging natural language expressions and standardized code identifiers. These candidates serve as initial references for verification against the Tabular List.

\textbf{Tabular List}. The Tabular List provides a structured listing of ICD codes, organized into chapters, categories, and subcategories. It includes code descriptions, inclusion terms, and exclusion notes that help coders verify candidate codes and select the appropriate level of specificity.

\textbf{Coding Guidelines}. The Coding Guidelines provide rules and conventions for applying ICD codes consistently, clarifying code usage and documentation requirements.

Together, these resources support clinical term lookup, code verification, and standardized code assignment.

\section{Scaling Law}

We apply SCL to models of different sizes, including Llama and Qwen families, to demonstrate the existence of scaling law.
As shown in Table \ref{table:scaling_law}, performance increases with model size.

\begin{table}[h]
\centering
\caption{\textbf{Scaling law of SCL under restricted code evaluation.}
Results on MDACE with Llama and Qwen backbones of different sizes.
P and R denote micro-averaged precision and recall.}
\label{table:scaling_law}
\tabsetup
\begin{tabular*}{\linewidth}{@{\hspace{\tabcolsep}\extracolsep{\fill}} l cccc}
\toprule
\textbf{Size} & \textbf{Micro-F1} & \textbf{Macro-F1} & \textbf{P} & \textbf{R} \\
\midrule
\rowcolor{tabgroup}
\multicolumn{5}{l}{\textbf{\textit{Llama-3.2}}} \\
1B & 40.8 & 16.0 & 48.6 & 35.2 \\
3B & 53.2 & 26.0 & 60.6 & 47.4 \\
\midrule
\rowcolor{tabgroup}
\multicolumn{5}{l}{\textbf{\textit{Qwen2.5}}} \\
0.5B & 31.8 & 10.6 & 44.2 & 24.9 \\
1.5B & 50.6 & 24.1 & 57.8 & 45.0 \\
7B   & 54.2 & 27.4 & 61.2 & 48.6 \\
\bottomrule
\end{tabular*}
\end{table}

\section{Implementation Details}
\label{sec:appendix_implementation}

For SCL, the document-level training data include 248 documents in total: 181 from MDACE and 67 from ACI, all drawn from their respective training splits.
Stage II uses an effective batch size of 16, comprising one complete document and 15 spans per batch.
Both stages are trained for 10 epochs. 
We train a single model once and evaluate the resulting checkpoint on both benchmarks.
For SFT, we use a batch size of 16 and train for 3 epochs, as longer training yielded no further improvements in our experiments.

For MDACE, where a single patient may have multiple notes, we concatenate all types of notes to form a single document, and use the union of codes as the ground truth. For ACI, we prompt LLMs to find supporting evidence for each code, in order to construct the corresponding evidence-grounded reasoning trace.

All the experiments are conducted on a single H20 GPU with 96GB of VRAM.
For LLM inference, we use vLLM \cite{vllm}.
For LLM fine-tuning, we use LLaMA-Factory \cite{llamafactory} with LoRA (rank = 8), a learning rate of 1e-4, and a cosine scheduler with a warmup ratio of 0.1.

\section{Case Study}
\label{sec:appendix_case_study}

\subsection{Comparing Different Paradigms}
\label{sec:appendix_paradigm}

\begin{figure*}[t]
    \centering
    \includegraphics[width=\textwidth]{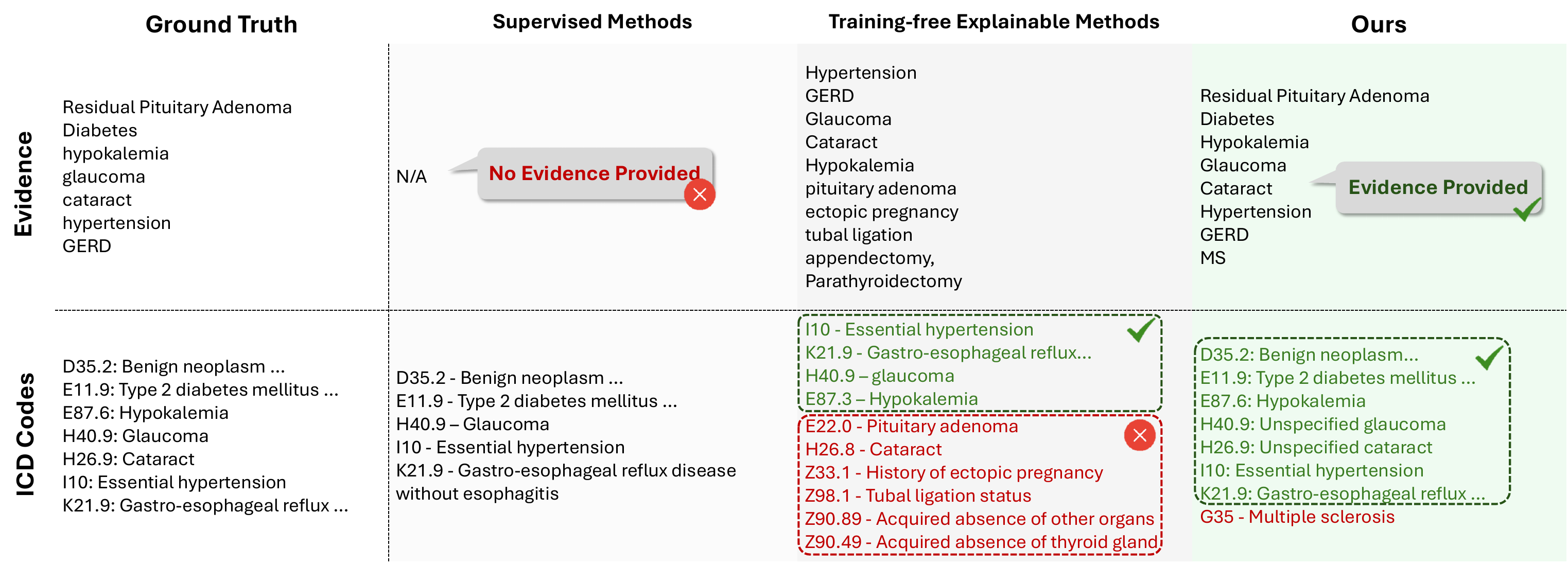}
    \caption{\textbf{Paradigm comparison on an MDACE test case.}
Supervised methods predict codes without evidence.
Explainable methods provide evidence but with lower coding accuracy.
SCL recovers both the evidence spans and the ICD codes that match the expert annotation.}
    \label{fig:case-study}
\end{figure*}

Figure~\ref{fig:case-study} compares the three paradigms on a case from the MDACE test set.
Traditional supervised methods map the note directly to ICD codes and give no evidence, so a coder cannot verify or revise the prediction.
Explainable methods can point to evidence, but with only a small set of evidence-annotated documents they miss or mis-assign codes.
SCL does both: it localizes the supporting spans and assigns the codes that match the expert annotation.
Span-centric supervision is what makes this possible, turning scalable span-level data into document-level coding that is accurate and inspectable.

\subsection{Failure Cases}
\label{sec:appendix_failure_case}

Disagreement with the benchmark is not always a model error.
Table~\ref{table:failure-case} shows one example from the ACI test set.

\begin{table*}[t]
\centering
\caption{\textbf{Failure case from the ACI test set.}
The note documents a right wrist sprain after the patient hit her arm on a pole while playing tennis.
SCL codes the injury, the mechanism, and the activity, while the gold annotation contains only the injury code.}
\label{table:failure-case}
\tabsetup
\newcommand{\ev}[1]{\textcolor{tabevid}{\textit{#1}}}
\begin{tabularx}{\textwidth}{>{\bfseries}p{1.8cm} >{\raggedright\arraybackslash}p{3.7cm} >{\ttfamily}l >{\raggedright\arraybackslash}X}
\toprule
Note & \multicolumn{3}{p{\dimexpr\textwidth-1.8cm-4\tabcolsep\relax}}{%
\textbf{CC:} Right arm pain. \newline
\textbf{HPI:} Ms.\ Martinez is a 43-year-old female who presents today for an evaluation of right arm pain after she \ev{hitting her arm on a pole} while \ev{playing tennis}. She had mild swelling at the time. There is pain with motion. She denies any numbness or tingling in her hand. She has a past medical history of allergies and takes Flonase. \newline
\textbf{Exam:} Examination of the left wrist shows tenderness over the distal radius and pain with flexion of the wrist and thumb. \newline
\textbf{Results:} X-rays of the left wrist show no obvious signs of acute fracture or other abnormality. \newline
\textbf{Impression:} \ev{Right wrist sprain}. \newline
\textbf{Plan:} At this point, I discussed the diagnosis and treatment options with the patient. I have recommended ice and anti-inflammatory medications. I am prescribing Motrin, 800 mg to take 3 times a day with food. She will follow up with me in 1 week or as needed.} \\
\midrule
Prediction & \ev{Right wrist sprain} & S63.501A & Unspecified sprain of right wrist, initial encounter \\
 & \ev{hitting her arm on a pole} & W22.8XXA & Striking against or struck by other objects, initial encounter \\
 & \ev{playing tennis} & Y93.73 & Activity, racquet and hand sports \\
\midrule
Gold & & S63.501A & Unspecified sprain of right wrist, initial encounter \\
\bottomrule
\end{tabularx}
\end{table*}

As shown in Table~\ref{table:failure-case}, SCL assigns \texttt{S63.501A} to the impression, matching the gold code.
It also assigns \texttt{W22.8XXA} to the documented mechanism and \texttt{Y93.73} to the documented activity.
ICD-10-CM guidelines use an external-cause code for how an injury occurred and an activity code from category \texttt{Y93} when the activity is stated.
Both facts are present in the note, so these two predictions follow the guidelines.
The benchmark lists only the injury code, and the cause codes are therefore counted as false positives.
This case shows that SCL may follow the coding guidelines and code the cause, while the benchmark itself is missing those codes.

\paragraph{Evidence rewriting errors.}
Generated evidence can also fail to map back to the source document when the model rewrites the original text.
Our inspection suggests that untraceable evidence mainly results from minor rewriting of the source text.
Table~\ref{table:evidence-rewriting-errors} shows five examples: a spelling error (\texttt{Macrosytosis}), a missing letter (\texttt{Astha}), a truncated word (\texttt{emia}), an inflection change (\texttt{fall} to \texttt{fell}), and abbreviation expansion (\texttt{vfib} to \texttt{ventricular fibrillation}).
Such changes can prevent lexical matching even when the output refers to the same clinical concept, motivating separate evaluation of semantic evidence matching and source traceability.

\begin{table}[t]
\centering
\caption{\textbf{Evidence rewriting errors that prevent source traceability.}
Model outputs and corresponding source text are reproduced verbatim; \textcolor{tabloss}{red} marks the rewritten parts.}
\label{table:evidence-rewriting-errors}
\tabsetup
\newcommand{\rw}[1]{\textcolor{tabloss}{#1}}
\begin{tabularx}{\linewidth}{>{\hsize=1.1\hsize\raggedright\arraybackslash}X >{\hsize=0.9\hsize\raggedright\arraybackslash}X}
\toprule
\textbf{Model output} & \textbf{Source text} \\
\midrule
\texttt{Macro\rw{s}ytosis} & \texttt{Macro\rw{c}ytosis} \\
\texttt{Astha} & \texttt{Asth\rw{m}a} \\
\texttt{emia} & \texttt{\rw{an}emia} \\
\texttt{f\rw{e}ll from \rw{a} ladder} & \texttt{f\rw{a}ll from ladder} \\
\texttt{\rw{ventricular fibrillation}} & \texttt{vtach/\rw{vfib} arrest} \\
\bottomrule
\end{tabularx}
\end{table}

\section{Evidence Extraction Statistics}
\label{sec:appendix_span_statistics}

Table~\ref{table:conflict} summarizes evidence availability when extracting spans from MIMIC-IV.
The extraction pipeline finds supporting spans for $78.2\%$ of code assignments, while $20.6\%$ yield \textit{No Evidence Found}. Some extraction outputs could not be parsed and are excluded from both categories.
Thus, for roughly one in five code assignments, no corresponding evidence is identified in the available discharge summary.

\begin{table}[t]
\centering
\caption{\textbf{Evidence availability in MIMIC-IV extraction.} Counts refer to code assignments. No evidence found denotes assignments for which the extraction pipeline identifies no supporting span in the available discharge summary. Some extraction outputs could not be parsed and are excluded from both outcome categories, so their counts do not sum to the initial total.}
\label{table:conflict}
\tabsetup
\begin{tabularx}{\linewidth}{l RR}
\toprule
\textbf{Category} & \textbf{Count} & \textbf{Unique codes} \\
\midrule
Initial assignments & 1,765,371 & 16,156 \\
\quad Supported & 1,380,071 & 14,275 \\
\quad No evidence found & 363,639 & 1,635 \\
\bottomrule
\end{tabularx}
\end{table}

\section{Span Quality Evaluation}
\label{sec:appendix_span_quality}

We randomly sample 1,000 codes, keeping the number of codes in each ICD-10-CM chapter as balanced as possible.
We retrieve the spans associated with these codes from the final span library, yielding 4,645 spans in total.
We conduct human evaluation of naturalness and plausibility on these spans and compare the judgments with LLM-based assessments.
Since all spans in the final library have passed both LLM-based criteria, agreement for each criterion is the proportion of spans that human evaluators also judge to satisfy it.

\begin{table}[t]
\centering
\caption{\textbf{Agreement between LLM and human evaluations.}
All 4,645 spans passed both LLM-based criteria. Matched/total reports the number also judged by human evaluators to satisfy each criterion over the total number evaluated.}
\label{table:appendix_span_quality}
\tabsetup
\begin{tabularx}{\linewidth}{l YY}
\toprule
\textbf{Criterion} & \textbf{Matched/total} & \textbf{Agreement (\%)} \\
\midrule
Naturalness  & 4,532 / 4,645 & 97.57 \\
Plausibility & 4,497 / 4,645 & 96.81 \\
\bottomrule
\end{tabularx}
\end{table}

Table~\ref{table:appendix_span_quality} reports agreement between human and LLM judgments.
Human evaluators judge 4,532 spans to satisfy naturalness and 4,497 to satisfy plausibility, corresponding to agreement rates of 97.57\% and 96.81\%, respectively.
These agreement rates are very high, particularly given that medical coding can involve ambiguous boundaries between related codes and that human judgments are not infallible.
The results support the use of LLM-based assessment for scalable span quality control within the final span library.
The remaining disagreements may reflect span quality issues, differences in interpretation, or human annotation errors; they should therefore not all be attributed to errors in the span library.
Future work could use expert adjudication of disputed spans to better characterize these disagreements and investigate the performance ceiling of our method under higher-quality span supervision.

\section{Prompts}
\label{sec:appendix_prompt}

\tcbset{
  promptcard/.style={
    enhanced,
    breakable,
    colback=white,
    colframe=teal!50!black,
    colbacktitle=teal!8,
    coltitle=black,
    fonttitle=\bfseries\small,
    fontupper=\footnotesize,
    boxrule=0.6pt,
    arc=1.5mm,
    left=2.4mm,
    right=2.4mm,
    top=1.2mm,
    bottom=1.4mm,
    toptitle=1.1mm,
    bottomtitle=1.1mm,
  }
}

\subsection{Prompts for SCL}

The prompts used for span-centric staged training are shown below.

\begin{tcolorbox}[promptcard, title={Document Data}]
Task:\\[0.35em]
You are a clinical coding assistant.\\[0.35em]
Your task is to analyze the provided clinical note, first extract all relevant clinical evidence spans that support diagnostic coding, and then output the corresponding ICD-10-CM codes.\\[0.7em]
\#\#\# Clinical Note:\\
\{text\}
\end{tcolorbox}

\begin{tcolorbox}[promptcard, title={Span Data}]
\#\#\# Evidence:\\[0.2em]
\{evidence\}\\[0.55em]
\#\#\# ICD-10-CM Codes:
\end{tcolorbox}

\subsection{Prompts for Interpretability Evaluation}
\label{sec:interpretability_prompt}

To evaluate model-predicted evidence, we use an LLM (GPT-5.1) to match LLM-predicted evidence spans with human-annotated evidence spans.
We then apply rule-based parsing to obtain the numbers of LLM-predicted evidence spans, human-annotated evidence spans, and matched pairs for computing the evaluation metrics.

\begin{tcolorbox}[promptcard, title={Predicted Evidence Evaluation}]
You are a clinical evidence evaluation expert.\\[0.35em]
You are given two unordered sets of clinical evidence spans.
Each line represents one evidence span.\\[0.35em]
1. Retain only meaningful clinical evidence spans.\\
\quad - Ignore empty lines, headings, or formatting text.\\
\quad - Semantically deduplicate each set, retaining one original span for each distinct finding.\\[0.35em]
2. Matching is semantic and lenient:\\
\quad - If a predicted span is more specific but clearly refers to the same clinical finding as a human span, count it as a match.\\
\quad - Minor wording differences do not matter.\\
\quad - Human annotations may be shorter or less specific.\\
\quad - If findings contradict (e.g., different laterality), do NOT count as a match.\\[0.35em]
3. Matching must be one-to-one.\\
\quad - One predicted span can match at most one human span.\\
\quad - One human span can match at most one predicted span.\\
\quad - Determine the optimal one-to-one matching that maximizes the number of matches.\\[0.35em]
4. Output only matched pairs, preserving the original span text. Write ``None'' if there are no matches.\\[0.7em]
\#\#\# Predicted Evidence:\\
\{evidence\}\\[0.45em]
\#\#\# Human-annotated Evidence:\\
\{human\_evidence\}\\[0.7em]
Output in markdown format exactly as:\\[0.3em]
- [predicted span] <-> [human-annotated span]
\end{tcolorbox}

\end{document}